%% file: main.tex
\documentclass[runningheads]{llncs}
\usepackage[T1]{fontenc}
\usepackage{graphicx}
\usepackage{cite}
\usepackage{amsmath}
\usepackage{amssymb}
\usepackage{amsfonts}
\usepackage{algorithmic}
\usepackage{graphicx}
\usepackage{textcomp}
\usepackage{xcolor}
\def\BibTeX{{\rm B\kern-.05em{\sc i\kern-.025em b}\kern-.08em
    T\kern-.1667em\lower.7ex\hbox{E}\kern-.125emX}}

\usepackage{algorithm}
\usepackage{algorithmic}
\usepackage{multirow}
\usepackage{dsfont}
\usepackage{bm}
\usepackage{makecell}
\usepackage{subcaption}
\usepackage{enumitem}
\usepackage{url}
\newcommand{\globalmodel}{$w$}

\newcommand{\localupdate}{$g_{n}$}
\usepackage{pifont}
\newcommand{\xmark}{\ding{55}}%
\usepackage{color,soul}
\usepackage{colortbl}
\definecolor{Gray}{gray}{0.9}
\usepackage{tikz}
\newcommand*\circled[1]{\tikz[baseline=(char.base)]{
            \node[shape=circle,draw,inner sep=0.2pt] (char) {#1};}}
\usepackage[capitalize]{cleveref}
\usepackage{array}
\newcommand{\PreserveBackslash}[1]{\let\temp=\\#1\let\\=\temp}
\newcolumntype{C}[1]{>{\PreserveBackslash\centering}p{#1}}
\newcommand{\revision}[1]{\textcolor{black}{#1}}
\newcommand{\camera}[1]{\textcolor{black}{#1}}

\begin{document}
\title{FLGuard: Byzantine-Robust Federated Learning via Ensemble of Contrastive Models}
\titlerunning{FLGuard: Byzantine-Robust FL via Ensemble of Contrastive Models}
%
\author
{Younghan Lee\inst{1,2}\and Yungi Cho\inst{1,2} \and Woorim Han\inst{1,2} \and Ho Bae \inst{3}\textsuperscript{*} \and Yunheung Paek\inst{1,2}\textsuperscript{*}}

\authorrunning{Y. Lee et al.}


\institute{ECE, Seoul National University, Seoul, Republic of Korea \and ISRC, Seoul National University, Seoul, Republic of Korea \and Dept. of Cyber Security, Ewha Woman's University, Seoul, Republic of Korea \\
\email{\{201younghanlee, q1w1ert1, rimwoo98, ypaek\}@snu.ac.kr, hobae@ewha.ac.kr}\\
{$^{*}$: Correspondence should be addressed to H. Bae and Y. Paek.}}
\maketitle              
\begin{abstract}
Federated Learning (FL) thrives in training a global model with numerous clients by only sharing the parameters of their local models trained with their private training datasets. Therefore, without revealing the private dataset, the clients can obtain a deep learning (DL) model with high performance. However, recent research proposed poisoning attacks that cause a catastrophic loss in the accuracy of the global model when adversaries, posed as benign \camera{clients}, are present in a group of clients. Therefore, recent studies suggested byzantine-robust FL methods that allow the server to train an accurate global model even with the adversaries present in the system. However, many existing methods require the knowledge of the number of malicious clients or the auxiliary (clean) dataset or the effectiveness reportedly decreased hugely when the private dataset was non-independently and identically distributed (non-IID). In this work, we propose FLGuard, a novel byzantine-robust FL method that detects malicious clients and discards malicious local updates by utilizing the contrastive learning technique, which showed a tremendous improvement as a self-supervised learning method. With contrastive models, we design FLGuard as an ensemble scheme to maximize the defensive capability. We evaluate FLGuard extensively under various poisoning attacks and compare the accuracy of the global model with existing byzantine-robust FL methods. FLGuard outperforms the state-of-the-art defense methods in most cases and shows drastic improvement, especially in non-IID settings. \url{https://github.com/201younghanlee/FLGuard}

\end{abstract}

\section{Introduction}
\input{Latex/1_introduction}

\section{Background}
\input{Latex/2_background}

\section{Threat Model \& Problem Setup}
\input{Latex/3_threatmodel}

\section{Our FLGuard} \label{sec:ourapproach}
\input{Latex/4_our_flguard}

\section{Evaluation}
\input{Latex/5_1_2_evaluation}
\input{Latex/5_3_result}

\section{Conclusion \& Future Work}
\input{Latex/6_conclusion}

\section{Acknowledgements}
This work was supported by Institute of Information \& communications Technology Planning \& Evaluation (IITP) grant funded by the Korea government (MSIT)
(No.2021-0-02068, Artificial Intelligence Innovation Hub,
No.RS-2022-00155966, Artificial Intelligence Convergence Innovation Human Resources Development, 
No.2022-0-00516, Derivation of a Differential Privacy Concept Applicable to National Statistics Data While Guaranteeing the Utility of Statistical Analysis (Ewha Woman's University and Seoul National University)
%
and the artificial intelligence semiconductor support program to nurture the best talents (IITP-2023-RS-2023-00256081),
Also, it was supported by
the BK21 FOUR program of the Education and Research Program for Future ICT Pioneers, Seoul National University in 2023 
and Inter-University Semiconductor Research Center (ISRC).

\appendix
\input{Latex/7_appendix}

\bibliographystyle{splncs04}
\bibliography{main}

\end{document}

%% file: Latex/1_introduction.tex
Federated learning (FL)~\cite{konevcny2016federated, mcmahan2017communication,kairouz2021advances} is an emerging distributed learning scheme in which the clients work collaboratively to train a high-performance model on the server (global model). Instead of sharing the client's private dataset, only the parameters of a model trained locally (local model) are uploaded to the server. In the era of big data, where the private dataset is considered highly valuable, protecting the privacy of the dataset is regulated by General Data Protection Regulation (GDPR)~\cite{EUdataregulations2018}. Therefore, many companies' \camera{service}, including Apple's Siri~\cite{https://doi.org/10.48550/arxiv.2102.08503}, Google's Gboard~\cite{googleaiblog_2017} employed FL schemes to provide their services in a privacy-preserving fashion. The clients are required to download the initial global model from the server. Using the private dataset, the local model update (gradient) is learned and then sent to the server for aggregation. For example, FedAvg~\cite{pmlr-v54-mcmahan17a}, one of the most commonly deployed FL schemes devised by Google, applies AGR that uses a weighted average of all local model updates from clients to form the global model update. Since the global model is trained using updates from all clients, the premise for FL to operate as designed is that all clients are legitimate and reliable.

However, according to~\cite{blanchard2017machine}, such a premise may no longer hold true due to a major threat to FL, called \textit{Byzantine failure}, that some clients can become adversaries whose aim is to thwart FL and degrade the quality of the global model by generating malicious local model updates. Such attacks, known as \textit{poisoning attacks}, can be categorized mainly into two types: model poisoning attack (MPA) and data poisoning attack (DPA).
The prerequisite of MPA is that the entire device of adversaries is compromised. The adversaries directly apply the perturbation vector ($p$) to the local model updates ($g$) so that the generated malicious model update ($g_m$) is given as follows: $g_m = g + \gamma p$, where $\gamma$ is the scale parameter. For example, LIE~\cite{baruch2019little} demonstrated that a very small $\gamma$ is sufficient to degrade the accuracy of a global model. More recently, MPAs exploited a new attack surface where the algorithm of AGR is known to  adversaries~\cite{fang2020local,shejwalkar2021manipulating}. Therefore, researchers converted the challenge of finding an appropriate $\gamma$ to an optimization problem. 
DPA assumes that adversaries only have access to private datasets. The adversaries poison the private dataset to hinder the training of the global model. The label-flip attack~\cite{fang2020local,shejwalkar2022back} is a typical DPA that intentionally changes the label of data to minimize the loss in the wrong direction.

To defend against both MPA and DPA, researchers proposed byzantine-robust FL schemes that filter out disguising malicious clients. Most of the proposed methods are \textit{statistic-based} using statistical information regarding the local updates to discard malicious clients.
For example, Trimmed-Mean (TM)~\cite{yin2018byzantine} introduced \textit{dimension-wise} filtering that treats each element in local model updates from all clients individually. 
Furthermore, Multi-Krum (MK)~\cite{blanchard2017machine} and DnC~\cite{shejwalkar2021manipulating} introduced \textit{vector-wise} filtering in which AGR decides to remove a certain number of clients.
SignGuard (SG)~\cite{xu2021signguard} uses the gradient statistics as features to filter out the malicious clients through clustering method.
FLtrust (FLT)~\cite{cao2020fltrust}, the \textit{validation-based} approach that utilizes the auxiliary dataset as a root-of-trust to validate the uploaded local model updates. 
%
However, existing defenses failed to achieve a high global model accuracy when the private dataset of clients was non-IID. Moreover, for the best results, current byzantine-robust FL schemes require prior knowledge regarding FL, such as the number of malicious clients or an auxiliary (clean) dataset for validation.

Due to the high cost of collecting human-annotated datasets, self-supervised learning~\cite{caron2020unsupervised, misra2020self}, such as contrastive learning (CL)~\cite{chen2020simple, khosla2020supervised, he2020momentum}, has erupted as a leading technique in the image classification domain. The main idea of CL is to contrast data against each other by pushing the data for the same representation close to each other and pulling apart that of different representations. 
Our insight is that the characteristic of CL, which congregates the data of a similar representation in the embedding space, is suitable for detecting the outlier among the local updates.
Hence, we adopt CL in FL security domain to establish a novel byzantine-robust FL by detecting and filtering malicious clients. 
In this paper, we propose FLGuard, a novel byzantine-robust FL scheme with contrastive models. Unlike the conventional CL, we additionally implement the dimension reduction technique to reduce the size of local updates in a tabular data shape. Also, we craft positive pairs by applying Gaussian noise to the local model updates. The rationale behind our approach is that a slight addition of perturbation does not change the true properties of original data in embedding space but rather induces the strong core properties shared between data to stand out and consequently to be extracted with ease. Only after projecting the data to the representations could we reveal the innate features of benign and malicious updates by contrasting data against each other.
Finally, we employ clustering methods only to use local updates from clients classified as benign in global model training.

We extensively evaluate FLGuard under various experimental settings. Empirically we demonstrate that FLGuard shows an impressive defensive performance even under non-IID settings without prior knowledge and thrives to achieve three defensive objectives: Fidelity, Robustness, and Efficiency. The high accuracy of the global model is maintained even without the presence of adversaries (fidelity). Especially, FLGuard achieved SOTA defense performance (robustness) under various types of poisoning attacks (DPA and MPA) and dataset distribution. Also, FLGuard does not incur a significant increase in communication cost on the server (efficiency).
Moreover, we performed ablation studies to confirm that FLGuard retained its SOTA performance regardless of the number of malicious clients and non-IID degree of the dataset. 
%
The following summarizes our contributions:
\begin{itemize}[wide]
\item We propose FLGuard, a novel byzantine-robust federated learning via an ensemble of contrastive models that operate without prior knowledge regarding the FL scheme (i.e., number of malicious clients, auxiliary dataset).
\item We conduct an extensive evaluation of FLGuard on various threat models, datasets, and poisoning attacks. The results show that FLGuard achieves state-of-the-art performance even in non-IID settings.
\end{itemize}

%% file: Latex/2_background.tex
\subsection{Federated Learning}\label{FL}
In federated learning (FL), the server trains a \textit{global model} ($w$) without disclosing the private dataset. We consider a widely used FL scenario with a server and $N$ clients with disjoint datasets~\cite{kairouz2021advances, konevcny2016federated, mcmahan2017communication}. We denote $D_n$ as $n$th client's private dataset.
In FL, the global model is obtained by solving the following optimization problem: $\min_{w} {F(w, D)}$ where $D = \cup_{n=1}^{N}{D_n}$ and $F$ is a loss function. Each client computes the local update (stochastic gradient), $g_{n} = \frac{\partial F(b,w)}{\partial w}$ over a mini-batch $b$ randomly sampled from $D_n$ and uploads it to the server synchronously. Finally, the server computes the global update ($G_{agr}$) using the local updates via an aggregation rule. 
Moreover, each client's private dataset ($D_n$) can be independently and identically distributed (IID) or non-independently and identically distributed (non-IID) among clients. Many federated learning aggregation rules struggle to achieve a high-performance global model under non-IID (real-world scenario) setting. We evaluate FLGuard under both \textit{IID} and \textit{non-IID} dataset settings among participating clients.

\subsection{Poisoning Attack}\label{PA}
According to~\cite{shejwalkar2022back}, two main factors that divide different types of poisoning attacks are adversaries' objectives and capabilities. The former refers to the intent of adversaries and a result that they can achieve. The latter depends on how much adversaries can thwart the federated learning given a certain level of access to the schemes. 
\textbf{Adversaries' objectives} can be categorized into two: discriminate or indiscriminate. Targeted attacks aim to induce the misclassification of specific sets of samples (discriminate), and untargeted attacks aim to misclassify any data samples (indiscriminate). 
\textbf{Adversaries' capabilities} represent either the adversaries have full access to the malicious client's \textit{model} or have access only to the \textit{data} of malicious clients. Model poisoning attacks (MPAs) have full access to the device of malicious clients and directly poison the local updates (model). Data poisoning attacks (DPAs) have only access to the local private dataset of malicious clients and poison the data samples (data). To prevent recent threats to FL, we focus and evaluate FLGuard on \textit{untargeted MPAs} and extend our attack scope to \textit{DPA}.

\subsection{Byzantine-robust Aggregation Rules}\label{BA}
Following~\cite{xu2021signguard}, we categorize the byzantine-robust aggregation rules into three different groups, which are discussed below.
\textbf{Statistic-based} defense strategies utilize statistical metrics to determine malicious clients.
Trimmed-Mean (TM)~\cite{yin2018byzantine} is a dimension-wise aggregation rule that uses the mean of the remaining local updates after removing \textit{m} largest and smallest values.
Multi-Krum (MK)~\cite{blanchard2017machine} adds the local update to the set until there are \textit{c} number of local updates such that $N - c > 2M +2$, where M is the upper bound of the number of malicious clients.
Bulyan (Bul)~\cite{guerraoui2018hidden} executes dimension-wise aggregation using TMean on the selected updates. 
Divide and Conquer (DnC)~\cite{shejwalkar2021manipulating} removes $e \cdot M$ local updates with the highest outlier score which is calculated using the singular value decomposition. 
SignGuard (SG)~\cite{xu2021signguard} uses a sign-based clustering method that incorporates gradient statistics as features.
\textbf{Validation-based} and current \textbf{learning-based} assume that the auxiliary dataset is available to the defender.
FLTrust (FLT)~\cite{cao2020fltrust} utilizes the auxiliary dataset as a root-of-trust and computes the trust score (TS) to reduce the impact of poisoning attacks from malicious clients.
Spectral anomaly detection~\cite{li2020learning} utilizes variational autoencoder (VAE) with clean auxiliary dataset to filter out the malicious local updates by measuring the reconstruction errors.
FLGuard belongs to \textit{vector-wise filtering, learning-based} byzantine-robust FL. Unlike previous byzantine-robust aggregation rules, FLGuard works \textit{without any prior information} (i.e., number of malicious clients or auxiliary dataset).

\subsection{Contrastive Learning and Clustering}\label{CL}
Recently, contrastive learning has shown promising results in self-supervised representation learning~\cite{chen2020big,he2020momentum}. The fundamental concept is to learning representations through maximizing agreement between augmented views of the same original image (positive pairs) and minimizing between that of the different images (negative pairs). SimCLR~\cite{chen2020simple} consists of a base encoder $f(\cdot)$ that extracts the representations $h$ from the original image $x$ (i.e., $f(x){=}h$) and a projection head $g(\cdot)$ that maps the representations $h$ to the latent vector $z$ (i.e., $g(h){=}z$). For the contrastive loss function~\cite{sohn2016improved}, normalized temperature-scaled cross-entropy loss ($NT$-$Xent$) is used with $x_i$, $x_j$ as positive pairs generated by randomly applying one of the augmentation operators. After training, only the base encoder $f(\cdot)$ is used for downstream tasks and a projection head $g(\cdot)$ is discarded. 
Agglomerative Hierarchical Clustering (AHC)~\cite{mullner2011modern} operates by grouping the data based on their similarities and the clusters with the shortest distance are grouped together repeatedly in a bottom-up manner until only a single cluster is formed. Then, the hierarchy tree (dendrogram) is dissected according to the desired number of clusters.
FLGuard employs \textit{contrastive models} based on SimCLR with self-supervised learning and uses \textit{clustering mechanism} to detect and discard malicious local updates.

%% file: Latex/3_threatmodel.tex
\begin{table}[!t]
    \centering
    \caption{Types of threat models based on adversaries' capability and knowledge}
    \input{Tables/1_threatmodel}
    \label{tbl:table1}
\end{table}

\textbf{Adversaries' objective}
Adversaries aim to poison the global model by devising malicious local updates and cause a considerable reduction in the accuracy. As discussed in the previous section, we focus on untargeted (indiscriminate) poisoning attacks.
\textbf{Adversaries' capability}
The adversaries are assumed to occupy \textit{M} malicious clients among a total of \textit{N} clients and can access the global model at each FL round. The number of malicious clients is assumed to be less than benign clients (i.e., $\frac{M}{N} < 0.5$)~\cite{shejwalkar2022back, xu2021signguard, shejwalkar2021manipulating, cao2020fltrust, fang2020local, zhao2022fedinv}. Adversaries with model poisoning capability are assumed to have full access to compromised clients, and they can directly manipulate the local updates. On the other hand, adversaries with data poisoning capability can only manipulate the local dataset. As shown in~\cref{tbl:table1}, threat models from Type-1 to Type-4 fall under model poisoning and Type-5 under data poisoning.
\textbf{Adversaries' knowledge}\label{knowledge} 
Following~\cite{shejwalkar2021manipulating}, we consider two aspects of adversaries' knowledge: local updates from benign clients and aggregation rule (AGR) algorithm from the server. There are a total of five different types of threat models, as shown in~\cref{tbl:table1}. Type-1 represents the \textit{strongest adversaries} with knowledge of both local updates of benign clients and AGR algorithm. While such type of threat model lacks practicality, previous works~\cite{fang2020local, baruch2019little} implemented such threat model to demonstrate the severe impact of poisoning attacks on FL. 
\textbf{Defender's objective}
The fundamental objective is to design FL scheme that attains byzantine robustness under MPA without compromising FL's fidelity and efficiency. Fidelity: Byzantine-robust FL scheme should not sacrifice the accuracy of the global model in return for the robustness when malicious clients are not involved (\cref{tbl:2_type_1_2}). Robustness: The defender's FL scheme should persist the accuracy of the global model even under the influence of poisoning attacks (\cref{tbl:2_type_1_2},~\cref{tbl:3_type_3_4}~\camera{and~\cref{tbl:4_type_5}).} Efficiency: Byzantine-robust FL scheme should not cause a significant overhead that will delay the training of the global model (\cref{fig:4_eff}).
\textbf{Defender's Capability} 
We assume the defense mechanism is operated from the server side and the server (defender) has an access to the global model and the local updates of all clients at each FL round. Unlike previous studies~\cite{shejwalkar2021manipulating,cao2020fltrust}, we further assume that the server does not have any prior knowledge regarding the malicious clients (i.e., the number of malicious clients) or access to the auxiliary dataset. The server is unaware of whether the local updates it receives are from malicious or benign clients.

%% file: Tables/1_threatmodel.tex
\begin{tabular}{|c|c|c|c|}
\hline
\multirow{3}{*}{Type}& 
\multirow{3}{*}{\makecell{Adversaries' \\ Capability}} &
\multicolumn{2}{c|}{\makecell{Adversaries' Knowledge}} \\
\cline{3-4}
&  &
\makecell{Local Updates of \\ Benign Clients} &
\makecell{Server's AGR \\Algorithm} \\

\hline
\hline

Type-1 (T1) & Model Poisoning & \checkmark & \checkmark \\
Type-2 (T2) & Model Poisoning  & \xmark & \checkmark \\
Type-3 (T3) & Model Poisoning & \checkmark & \xmark \\
Type-4 (T4) & Model Poisoning & \xmark & \xmark \\
Type-5 (T5) & Data Poisoning & \xmark & \xmark \\

\hline
\end{tabular}

%% file: Latex/4_our_flguard.tex
\subsection{Overview}\label{IV-A}
\begin{figure}[!t]
  \centering
  \includegraphics[width=0.7\columnwidth]{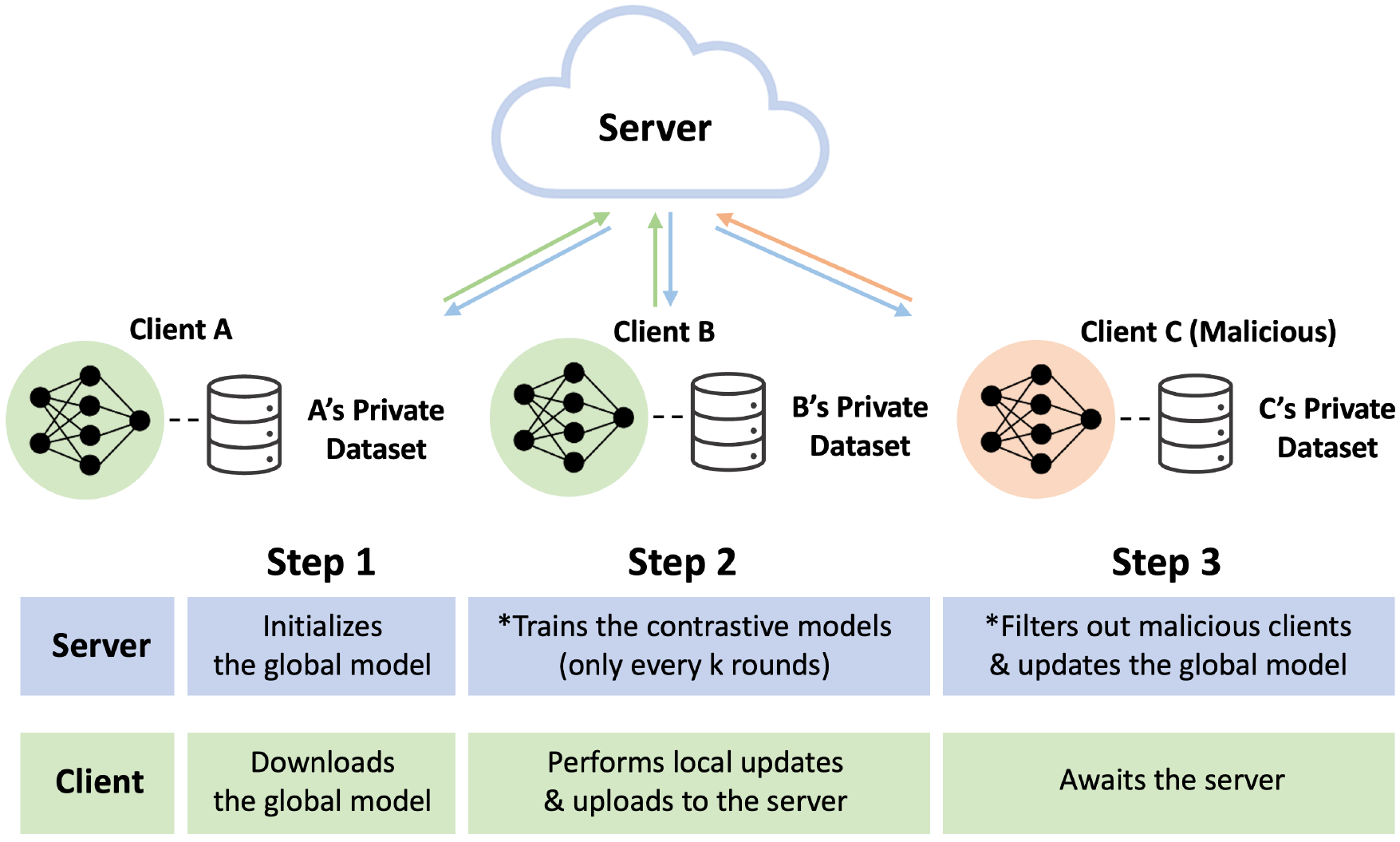}
\caption{Overview of byzantine-robust FL. Steps described in asterisk (*) denote the design novelty of our FLGuard}
\label{fig:1_FL_overview}
\end{figure}

\cref{fig:1_FL_overview} illustrates the overview of byzantine-robust FL where a single malicious client is present. FL operates essentially in three steps, both from the server and client sides. Explanations with asterisks denote the part in which FLGuard is different from plain FL without a defense mechanism against adversaries.

\begin{algorithm}[!ht]
  \caption{FLGuard}  \hspace*{\algorithmicindent} \textbf{Intput:} Total number of clients $N$, number of FL rounds $R$, number of local iterations $I$, loss function $F$, private dataset $D$, batch size $b$, global learning rate $\eta$, local learning rate $\alpha$, FLGuard parameter $k$
  
  \hspace*{\algorithmicindent} \textbf{Output:} Global model $w_R$
  
\begin{algorithmic}[1]
    \label{alg:FLGuard}
    \STATE $w_0 \gets$ initialize at random
        \FOR{each FL round $r = 1,2,...,R$}
        \FOR{each client $n = 1,2,...,N$ \textbf{in parallel}}
            \STATE $g_n \gets$ \textbf{LocalUpdate}($w_{r-1}$, $I$, $D_n$, $b_n$, $F$, $\alpha$)
        \ENDFOR
        \STATE $G \gets$ set of current local updates $g_1, g_2, ..., g_N$ 
        \STATE $G_{train} \gets $ set of local updates for last k FL rounds
               \IF{$r$ $mod$ $k = 0$}
        \STATE $ContrastiveModels, Scaler \gets \textbf{Training} (G_{train})$
        \ENDIF
        \STATE $G_{lv},\space G_{rd} \gets \textbf{Preprocessing}(G,Scaler)$
        \STATE $C_{good} \gets  \textbf{Filtering} (G_{lv},G_{rd}, ContrastiveModels)$ 
        \STATE $G_{filtered} \gets G[C_{good}]$
        \STATE $G_{agr} \gets \frac{1}{|C_{good}|} \cdot \sum G_{filtered}$
        \STATE $w_{r} = w_{r-1} - \eta \cdot G_{agr}$ 
    \ENDFOR
\end{algorithmic}
\end{algorithm}

\textbf{Step 1:}
\textit{Server Side}: The server initializes the global model (\globalmodel) at random in the first round of FLGuard (Line 1 in~\cref{alg:FLGuard}). The global model is off-loaded to all clients in FL for synchronization.
\textit{Client Side}: The clients download the global model (\globalmodel) from the server (Line 1 in~\cref{alg:LocalUpdate}).
\textbf{Step 2:}
\textit{Server Side}: Unlike other FL schemes in which the server awaits the clients to upload the local updates, FLGuard capitalizes on such an idle state to train the contrastive models. FLGuard only trains the new model every \textit{k} rounds of FL as the same contrastive models can be used for multiple FL rounds (Line 7-10 in~\cref{alg:FLGuard}). Further details on the training phase are explained in~\cref{IV-C}.
\textit{Client Side}: The clients perform local updates (\localupdate) on the global model with mini-batch sampled from their private dataset ($D_n$). All clients perform in parallel and send calculated local updates to the server (Line 2-6 in~\cref{alg:LocalUpdate} in~\cref{appendixA}).
\textbf{Step 3:} 
\textit{Server Side}: In the final step, the server filters out malicious local updates and removes them from updating the global model (\cref{alg:Filtering} in ~\cref{appendixA}). The details of each function in the filtering phase of FLGuard are explained in~\cref{IV-D}. Finally, the server aggregates the filtered local updates by averaging to form the global update ($G_{agr}$) (Line 12-15 in~\cref{alg:FLGuard}).~\textit{Client Side}: The clients await until the next round of FL while the server updates the global model.

\begin{figure*}[!t]
  \centering
  \includegraphics[width=\textwidth]{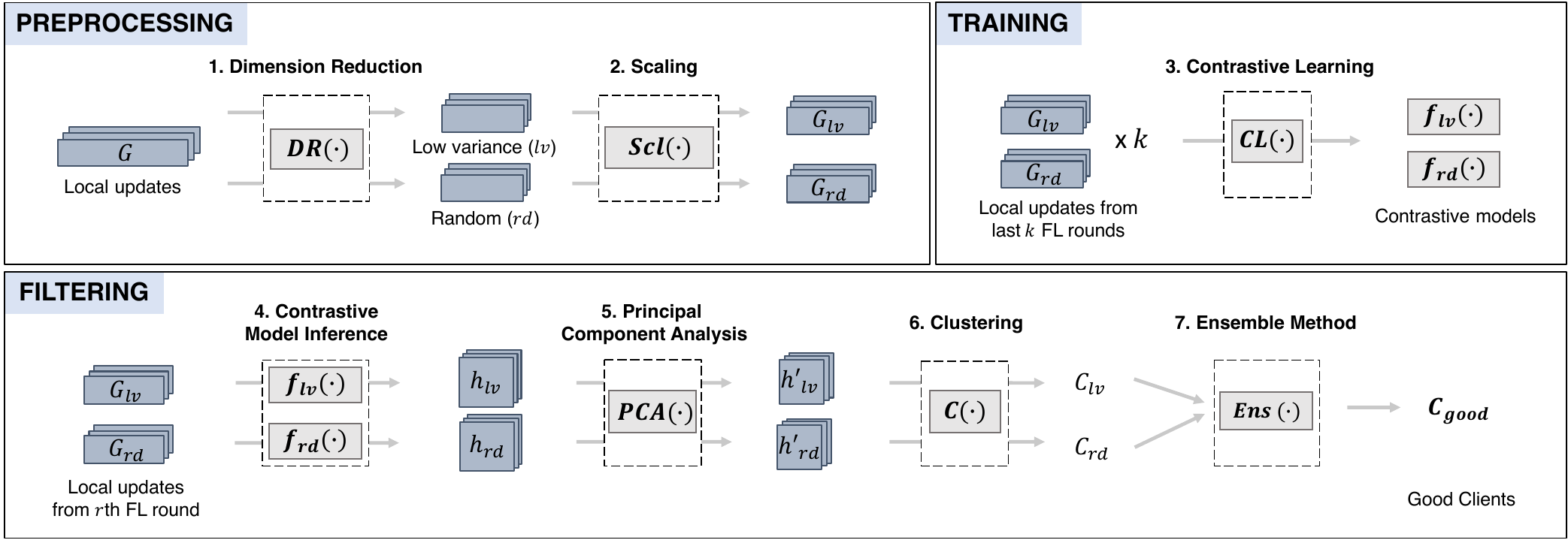}
\caption{Flowchart of FLGuard Process (Preprocessing, Training Contrastive Models, and Filtering Malicious Clients). We note that the training phase only occurs every \textit{k} FL rounds to generate new contrastive models}
\label{fig:2_flowchart}
\end{figure*}

\cref{fig:2_flowchart} summarizes the process of FLGuard, which consists of three parts: preprocessing, training, and filtering. In the preprocessing phase, we prepare local updates by applying popular data preprocessing techniques. In the training phase, we utilize contrastive learning to generate contrastive models that extract local updates' representation. In the filtering phase, through the clustering and ensemble method, FLGuard only collects benign clients ($C_{good}$).

\subsection{Preprocessing Local Updates}
\subsubsection{Dimension Reduction and Scaling} 
Previous study reported that the robustness of FL is compromised with high dimension (i.e., \textit{the curse of dimensionality}~\cite{shejwalkar2021manipulating}) and utilized a random filter. However, we additionally implement a low variance filter in FLGuard that removes features with low variance to avoid dismissing the elements that represent the unique aspect of local updates depending on the origin (malicious or benign clients). Given a set of local updates of clients ($G$), we compute variance (i.e., $\sigma^2 {=} \sum (x_i {-} \mu)^2 / N$ where $N{=}$ number of clients) of each element among local updates and remove the element with low variance to fit the dimension of 3,072. Scaling the data is necessary to prevent the data with a large magnitude from dominating the data with a small magnitude. As the values of local updates are unbounded in magnitude, an appropriate scaling technique must be applied. We observed that the local updates are either positive or negative. Therefore, we implement a max-absolute (MaxAbs) scaler to maintain the sign of local updates. MaxAbs ($x_{scaled}{=}\frac{x}{max(\lvert x \rvert)}$) scales the data to a range from -1 to 1. As shown in~\cref{fig:2_flowchart}, $G_{lv}$ and $G_{rd}$ are the final results of preprocessing phase.

\subsection{Training Contrastive Models} \label{IV-C}
\subsubsection{Contrastive Learning}
\begin{figure}[!t]
  \centering
  \includegraphics[width=0.5\columnwidth]{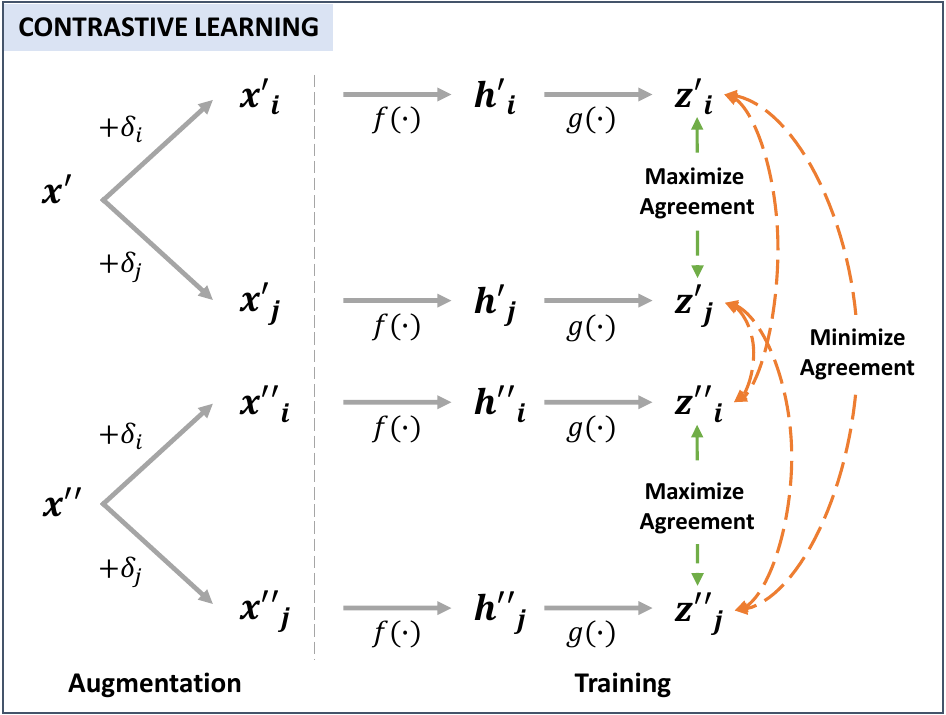}
\caption{Illustration of Contrastive Learning including Augmentation. $f(\cdot)$ and $g(\cdot)$ represent the contrastive models (encoders) and the projection heads}
\label{fig:3_contrastive_learning}
\end{figure}

Collecting a vast labeled dataset for training is expensive as human supervision is required. Especially in the FL scenario, dealing with high-dimensional local updates as the training dataset makes collecting the annotated dataset seems unfeasible. Therefore, we utilize self-supervised learning, which automatically labels the dataset by learning the difference in representations of inputs.
In the training phase, we adopt contrastive learning, which enforces the data of similar representations closer and different representations further away by maximizing the agreement of positive pairs and minimizing the agreement of negative pairs.
Our implementation of contrastive learning is based on SimCLR~\cite{chen2020simple} with modifications to suit the security domain. The main difference is that the local updates are tabular data.
Augmentation strategy, applied to the original data to form positive and negative pairs, is vital for learning good representations of the data. \revision{Inspired by~\cite{NEURIPS2021_9c8661be,bahri2021scarf}, we choose to add Gaussian noise ($X \sim \mathcal{N}(\mu,\sigma^{2})$) by using a masking ratio as means of augmenting the original tabular data (local updates). We generate two augmented data (views) per each original data by randomly adding Gaussian noise with $\sigma^{2} = 0.01$ and maksing ratio of 0.1.} The augmented views obtained from the same original data are treated as positive pairs (i.e., $x'_i \leftrightarrow x'_j$ and $x''_i \leftrightarrow x''_j$ in~\cref{fig:3_contrastive_learning}). The negative pairs are simply formed by pairing each data with all other data in the batch except for itself. The orange dotted line in~\cref{fig:3_contrastive_learning} represents the negative pairs. For example, with 32 as the batch size ($B$), we create a total of 32 positive pairs and 1,984 negative pairs~\camera{(i.e., 2($B$-1) negative pairs per one sample).}

Training the contrastive models $f(\cdot)$ and the projection heads $g(\cdot)$ require the pairs of augmented views and the loss function (\textit{NT-Xent}) applied to the space mapped by the projection heads can be written as below~\cref{eqn:one}.
\begin{equation} \label{eqn:one}
    l(z_{i,j}) = 
    -\mathrm{log}\frac{\mathrm{exp}(sim(z_{i,j})/ \tau)}{\sum^{2B}_{k=1} \mathds{1}_{[k \neq i]} \mathrm{exp}(sim(z_{i,k})/\tau)}
\end{equation}
\begin{equation} \label{eqn:two}
    L = \frac{1}{2B} \sum_{i,j=1}^{B}[l(z_{i,j}) + l(z_{j,i})] 
\end{equation}
    where $\mathds{1}_{[k \neq i]}$ is an indicator function that outputs 1 if and only if $k \neq i$ and $\tau$ is a temperature parameter.

The final loss ($L$) is computed by averaging over all pairs in the batch as shown in~\cref{eqn:two}. \revision{We use the dot product between l2 normalized two vectors (i.e., cosine similarity) that given two vectors \textit{a} and \textit{b}, $sim(a,b) = a^Tb / \lVert a \rVert_{2} \cdot \lVert b \rVert_{2}$ as the similarity function and set $\tau=0.01$.} Since the local updates are tabular data, we design the architecture of contrastive models $f(\cdot)$ by following~\cite{NEURIPS2021_9c8661be} with a single layer encoder with leaky-Rectified Linear Unit (leaky-ReLU) as activation function ($\mathds{1}_{[x<0]}\alpha x+\mathds{1}_{[x>=0]}x$ where $\mathds{1}$ is an indicator function, and $\alpha$ is a small constant).
The projection head $g(\cdot)$ is a 2-layer multi-layer perceptron (MLP) which is discarded once the contrastive models are trained due to the information lost from the contrastive loss. Note that the last layers of both the encoder and projection head are linear, with a dimension of 3,072. Finally, as shown in~\cref{fig:2_flowchart}, we create two contrastive models $f_{lv}(\cdot)$ and $f_{rd}(\cdot)$ depending on the results of filters used in preprocessing.

\subsubsection{Contrastive Model Training Interval}
As the values of local updates vary throughout FL rounds, employing the same contrastive models trained with earlier rounds to filter out malicious local updates from the later rounds would not guarantee the best performance. Therefore, we update our contrastive models in FLGuard every \textit{k} FL rounds where \textit{k} is a parameter to adjust the interval of updating contrastive models. From a certain round in FL, we use local updates from the last \textit{k} FL rounds (e.g., from 1 to 5) to train the contrastive models which are used for the next \textit{k} rounds (e.g., from 6 to 10) to filter the malicious clients. For example, for 1,500 total FL rounds, updating new contrastive models occurs 300 times when \textit{k} is set to 5 (i.e., every 5 FL rounds).

\subsection{Filtering Malicious Clients}\label{IV-D}
In the filtering phase of FLGuard, we utilize both contrastive models only to collect local updates from good clients (i.e., $C_{good}$). More specifically, we apply principal component analysis (PCA) and clustering algorithms on the representations of local updates to filter out malicious clients. 
\textbf{Contrastive Model Inference}
As shown in~\cref{fig:2_flowchart}, in each FL round, we project the local updates $G_{lv}$ and $G_{rd}$ to representation $h_{lv}$ and $h_{rd}$ with the contrastive models $f(\cdot)$. We only use the encoder trained in the training phase for the inference and discard the projection head. 
\textbf{Principal Component Analysis (PCA)} \revision{For dimension reduction process, we empirically set on the number of components equal to two and we obtained dimension-reduced $h'_{rd}$ and $h'_{lv}$ as the results.}
\textbf{Clustering}
We form two clusters with the latent space representations and assume that the larger is from benign clients. The results of clustering are groups of benign clients ($C_{lv}$ and $C_{rd}$) depending on different representations ($h'_{lv}$ and $h'_{rd}$). In this paper, we design FLGuard using agglomerative hierarchical clustering (AHC) with single linkage metric to measure the distance between the clusters. 
\textbf{Ensemble Method}
To maximize the robustness of FLGuard, we implement the ensemble method with two different clusters which are formed by two different contrastive models. We only select clients that are classified as benign by both contrastive models (i.e., $C_{good} = C_{rd} \cap C_{lv}$). Finally, the local updates uploaded by clients in $C_{good}$ are used to train the global model.

%% file: Latex/5_1_2_evaluation.tex
\subsection{Experimental Setup}
\subsubsection{Datasets}
To distribute the training data among participating clients~\cite{cao2020fltrust}, we divide the clients into \textit{n} groups (i.e., the number of classes in a dataset). Then, a sample from the dataset with label \textit{K} is assigned to group \textit{K} with probability \textit{q} and to any other group with $\frac{1-\textit{q}}{\textit{n}-1}$. Finally, samples in the same group are uniformly distributed to each client. We note that the probability (\textit{q}) is the parameter that determines the distribution of the private dataset among clients. For example, if \textit{q} = $\frac{1}{\textit{n}}$, the private training dataset are independent and identically distributed (IID), otherwise non-IID. 
The greater the value of \textit{q}, the higher degree of non-IID.~\circled{1} MNIST~\cite{lecun1998gradient} contains gray-scale digit images of 10 classes. MNIST-0.1 ($q{=}0.1$) indicates IID and MNIST-0.5 ($q{=}0.5$) simulates non-IID.~\circled{2} CIFAR-10~\cite{cifar} is a 10-class RGB image classification dataset and the training data are IID among participating clients.~\circled{3} FEMNIST~\cite{caldas2018leaf} is a 62-class character recognition classification dataset consisting of 3,400 clients and each client has a different number of images (non-IID). For FL training, we randomly select 60 clients in each round to participate~\cite{shejwalkar2021manipulating}. We set the percentage of malicious clients in each round to be 20\% of total clients~\cite{cao2020fltrust,shejwalkar2021manipulating,xu2021signguard}. The details of FL parameter setting for each dataset can be found in~\cref{appendixA}.

\subsection{Evaluated Poisoning Attacks}
A general approach to model poisoning attacks (MPAs) is as the following. Firstly, adversaries compute an average of available benign local updates ($g_b$). The availability depends on the type of threat model as explained in~\cref{knowledge}. After selecting an appropriate perturbation vector ($p$), the scale ($\gamma$) is multiplied before being added to $g_b$. Finally, the malicious clients upload $g_m = g_b + \gamma p$ to the server. In this paper, following~\cite{shejwalkar2021manipulating}, we implement two types of perturbation vectors ($p$):
\circled{1} Inverse unit vector ($\bm{uv} = -\frac{g_b}{\left\lVert g_{b} \right\rVert_{2}}$)
\circled{2} Inverse sign ($sgn = - \textrm{sgn}(f_{avg}(g))$). 
We evaluate FLGuard against various MPAs including optimization approaches~\cite{fang2020local,shejwalkar2021manipulating} and adaptive attacks~\cite{shejwalkar2021manipulating} where AGR is known to the adversaries (i.e., threat model T1 and T2). In particular, we design an adaptive attack on FLGuard to evaluate the robustness under a severe attack in which the adversaries can obtain the contrastive models trained. In terms of perturbation vector, we set $p{=}sgn$ for optimization approaches and adaptive attacks.
Moreover, when AGR is unknown to the adversaries (i.e., threat model T3 and T4), we evaluate FLGuard against Little Is Enough (LIE)~\cite{baruch2019little}, Min-Max and Min-Sum~\cite{shejwalkar2021manipulating}, and Sign-Flip Attack (SF)~\cite{rajput2019detox}.
Also, Label-Flip attack~\cite{fang2020local,cao2020fltrust,shejwalkar2022back} is one of the data poisoning attacks. The malicious local updates are generated using the local private dataset with flipped labels from malicious clients.
The details of each evaluated attacks can be found in~\cref{appendixB}.

%% file: Latex/5_3_result.tex
\subsection{Experimental Result} \label{result}
\subsubsection{Fidelity}\label{fidelity}
In order for byzantine-robust FL to be suitable in a real-world scenario, the fidelity of FL scheme must not be violated when no adversaries are present. \camera{The abbreviation for AGR can be found in~\Cref{BA}.} We evaluated the fidelity of FLGuard (FLG) by comparing the global model accuracy with other AGRs and consider FedAvg as the baseline. FedAvg achieved 97.24\%, 73.54\%, 97.16\%, and 84.11\% for MNIST-0.1, CIFAR-10, MNIST-0.5, and FEMNIST respectively. As shown in \textit{No Attack} column in~\cref{tbl:2_type_1_2}, the accuracy trained by FLGuard is higher than other AGRs and as high as FedAvg. For all four datasets, the accuracy of the global model trained by FLGuard only drops by less than 1\% compared to that of FedAvg. Especially in MNIST-0.1, FLGuard did not induce any drop in accuracy and in FEMNIST, FLGuard managed to achieved even higher accuracy than FedAvg as all malicious clients were filtered out \camera{(84.11\% for FedAvg and 84.74\% for FLGuard)}. In CIFAR-10, the accuracy drops was only 0.81\% for FLGuard while DnC, FLTrust, and SignGuard drop by 1.10\%, 2.80\%, and 2.90\%. FLGuard outperforms \camera{others} because the filtering performance is better compared to DnC and SignGuard and FLTrust operates in dimension-wise manner that considers all local updates to train the global model. Therefore, we conclude that the fidelity of FLGuard is achieved for all four datasets.

\begin{table*}[!ht]
    \centering
    \caption{Comparison of accuracy from byzantine-robust AGRs against no attack (Fidelity) and MPAs (Robustness) under threat model Type-1 \& Type-2}
    \input{Tables/2_type_1_2}
    \label{tbl:2_type_1_2}
\end{table*}

\begin{table*}[!ht]
    \centering
    \caption{Comparison of global model accuracy from byzantine-robust AGRs against and MPAs (Robustness) under threat model Type-3 \& Type-4}
    \input{Tables/3_type_3_4}
    \label{tbl:3_type_3_4}
\end{table*}

\subsubsection{Robustness}
FLGuard has been extensively tested under many threat models with various poisoning attacks including MPAs (\cref{tbl:2_type_1_2} and \cref{tbl:3_type_3_4}) and DPA (\cref{tbl:4_type_5}). The robustness of FLGuard is analyzed based on three criteria: types of threat model, dataset distribution, and types of attacks.
\textbf{Robust under different types of threat models}.
Among 24 cases in Type-1 (i.e., the strongest threat model) FLGuard achieved the best performance in 18 cases (75\%). 
In Type-2, where adversaries are unaware of local updates of benign clients, FLGuard outperformed in 87.5\% (21/24) of total cases.
We concluded that FLGuard is robust to MPAs even under strongest adversaries and outperforms other defenses by up to 79.5\% (vs. FLT in FEMNIST).
In Type-3 (i.e., unknown AGR algorithm), FLGuard outperformed other defenses in 71\% (17/24) of total cases
and in Type-4 (i.e., weakest adversaries) FLGuard achieved the best accuracy in 67\% of cases (21/28). In particular, FLGuard improved the global model accuracy up to 78.45\% in Type-4 (vs. FLT in FEMNIST).
Finally, we expand the evaluation to DPA (Type-5) by executing label-flip attacks (SLF and DLF).~\cref{tbl:4_type_5} shows FLGuard achieves state-of-the-art performance in 5 out of 8 cases.
\textbf{Robust under both IID and non-IID}. 
For IID dataset (MNIST-0.1 and CIFAR-10), FLGuard achieved the best accuracy in 76\% of all cases (41/54). 
Additionally, in non-IID settings where many byzantine-robust FL are known to fail in achieving a high accuracy global model, FLGuard also achieved the best accuracy in 76\% of all case (41/54). 
In particular, we note that FLGuard achieved impressive results in FEMNIST. For example, against Min-Max ($sgn$), FLGuard achieved 82.63\% compared to 5.25\%, 6.17\%, and 8.13\% in previous defenses (DnC, FLTrust, and SignGuard). Furthermore, FLGuard demonstrated a notable absence of catastrophic failure in comparison to other defense mechanisms. For instance, when examining FEMNIST dataset, FLGuard exhibited mean and standard deviation values of 83.58 and 0.89, respectively, whereas Signguard yielded values of 62.68 and 32.54.
\textbf{Robust against various types of poisoning attacks}. 
As explained above, we evaluated AGRs against both MPAs and DPAs. While FLGuard exhibited its robustness across all types of poisoning attacks, the defense against optimization and adaptive attacks (best in 39/48 cases) and Min-Sum (14/16) attacks was particularly outstanding. However, FLGuard achieved relatively lower accuracy against LIE attack (8/16).

\subsubsection{Efficiency}\label{efficiency}
\begin{figure}[!ht]
\centering 
    \begin{subfigure}[b]{0.24\columnwidth}
    \includegraphics[width=\columnwidth ]{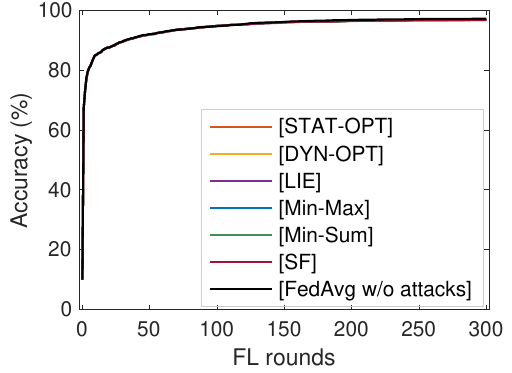}
    \subcaption{MNIST-0.1}
    \end{subfigure}
    \begin{subfigure}[b]{0.24\columnwidth}
    \includegraphics[width=\columnwidth ]{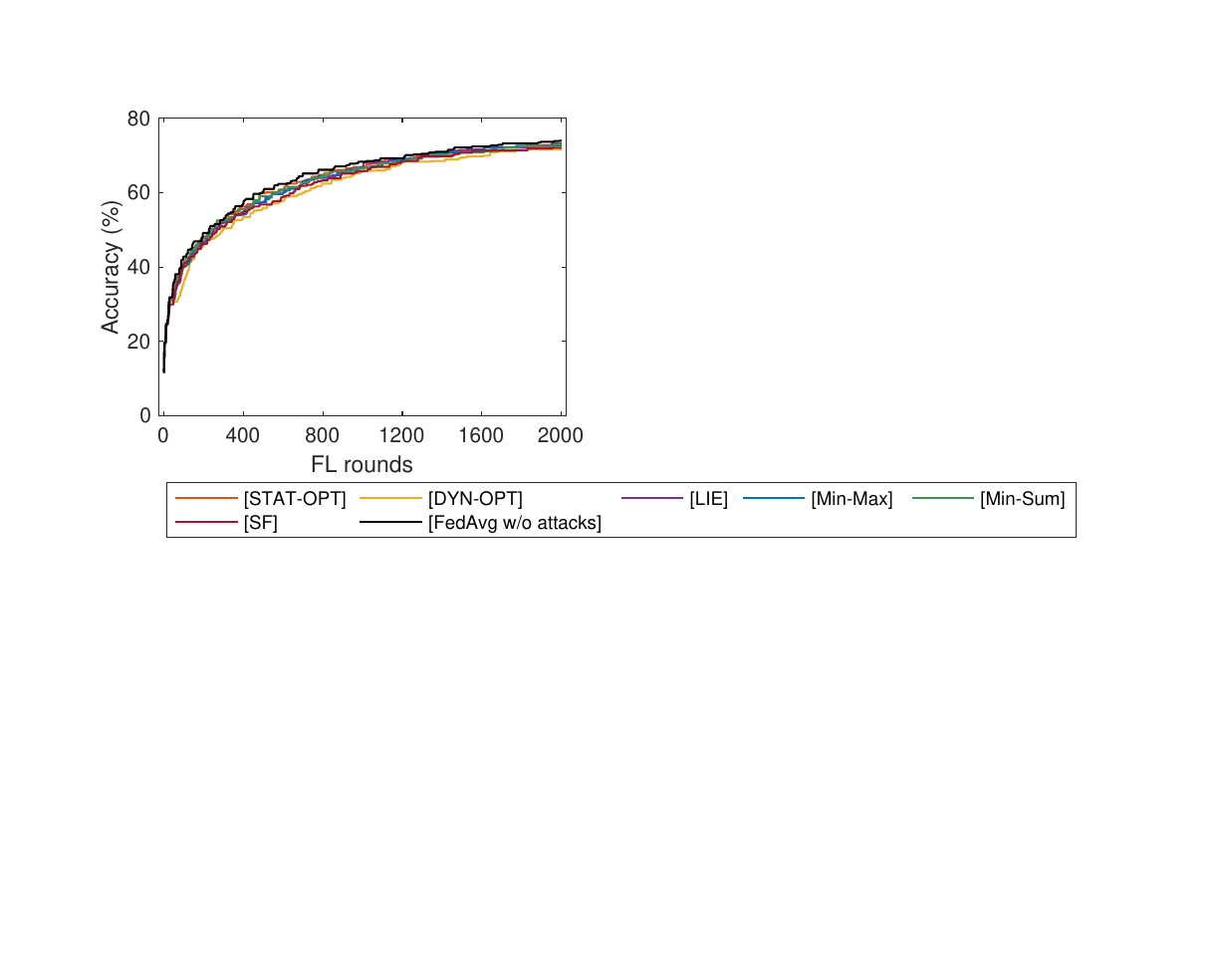}
    \subcaption{CIFAR-10}
    \end{subfigure}
    \begin{subfigure}[b]{0.24\columnwidth}
    \includegraphics[width=\columnwidth ]{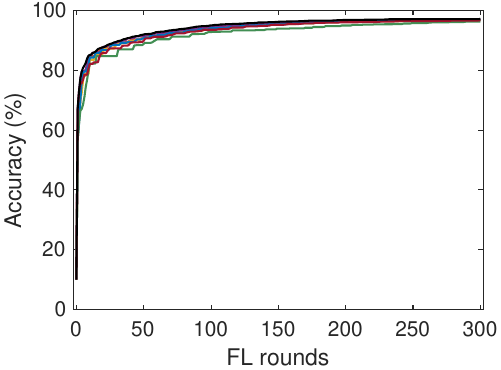}
    \subcaption{MNIST-0.5}
    \end{subfigure}
    \begin{subfigure}[b]{0.24\columnwidth}
    \includegraphics[width=\columnwidth ]{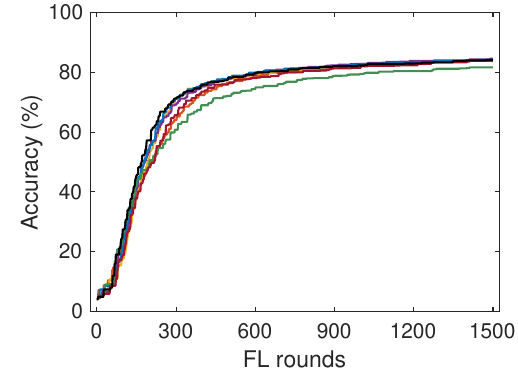}
    \subcaption{FEMNIST}
    \end{subfigure}
\caption{Global Model Accuracy vs. FL rounds for FLGuard with various MPAs and FedAvg without attacks}
\label{fig:4_eff}
\end{figure}

We discuss efficiency by comparing the number of rounds and the total time required to train the global model~\cite{cao2020fltrust}. The defender aims to implement byzantine-robust AGR without incurring extra overhead. As shown in~\cref{fig:4_eff}, FLGuard does not pose extra communication costs (extra FL rounds) for the clients since the accuracy of the global model under attacks saturates as fast as FedAvg without poisoning attacks. We also measured the time taken to train the contrastive models and filter out the malicious clients with a Linux machine running on a Xeon Gold 6226R with 16 cores, 256GB of RAM, and NVIDIA RTX A6000 GPU. For FEMNIST, given $k{=}5$, the training time was 22.1s, and the filtering time was 78ms. As FLGuard capitalizes the idle time on the server (waiting for the clients to upload the local update) to train the contrastive models, the extra overhead is minimal given a suitable computational power.

\begin{table}[!t]
    \centering
    \caption{Comparison of global model accuracy from byzantine-robust AGRs against DPAs under threat model Type-5}
    \input{Tables/4_type_5}
    \label{tbl:4_type_5}
\end{table}

\subsection{Ablation Study} \label{abal}
\subsubsection{Impact of Number of Malicious Clients}
Sub-figures of (a) to (d) (i.e., first row) in~\cref{fig:5_ablation} depict the impact of different fractions of malicious clients present in FL that varies from 0\% to 40\% which is an extreme adversarial setting~\cite{xu2021signguard,cao2020fltrust,shejwalkar2021manipulating}. The performance of the global model trained with other defense mechanisms deteriorates as the proportion of malicious clients increases. This effect is particularly pronounced when the percentage of malicious clients is high (40\%), where the decline in accuracy is significant for T1 and T2 attacks. In contrast, FLGuard maintains its robustness even as the proportion of malicious clients increases across all types of attacks. Notably, when facing the Min-Sum attack, only FLGuard managed to keep the accuracy close to that of FedAvg without attacks, while other FL methods saw a sharp decline to less than 10\%.

\subsubsection{Impact of Non-IID Degree}
\begin{figure*}[!t]
    \begin{subfigure}[b]{0.24\textwidth}
    \centering
    \includegraphics[width=\columnwidth]{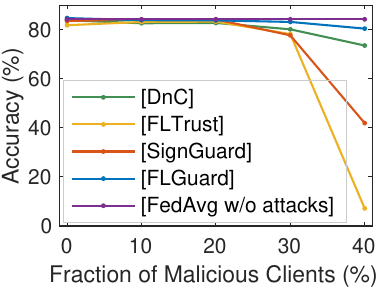}
    \subcaption{T1 (STAT-OPT)}
    \end{subfigure}  
    \begin{subfigure}[b]{0.24\textwidth}
    \centering
    \includegraphics[width=\columnwidth]{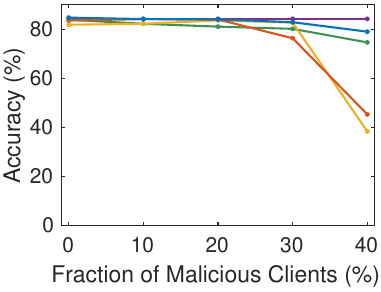}
    \subcaption{T2 (STAT-OPT)}
    \end{subfigure} 
    \begin{subfigure}[b]{0.24\textwidth}
    \centering
    \includegraphics[width=\columnwidth]{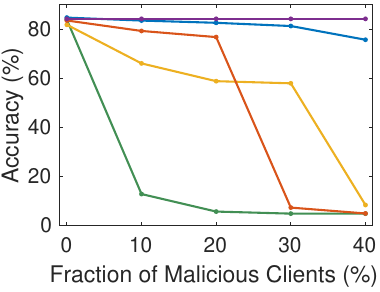}
    \subcaption{T3 (Min-Sum)}
    \end{subfigure} 
    \begin{subfigure}[b]{0.24\textwidth}
    \includegraphics[width=\columnwidth]{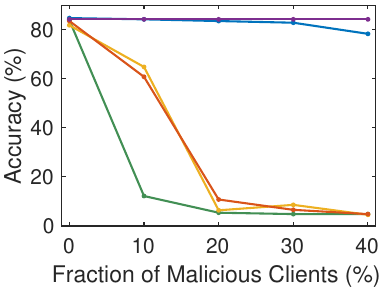}
    \subcaption{T4 (Min-Sum)}
    \end{subfigure}  
    
\begin{subfigure}[b]{0.24\textwidth}
    \centering
    \includegraphics[width=\columnwidth]{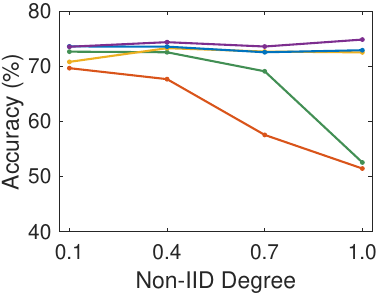}
    \subcaption{T1 (DYN-OPT)}
    \end{subfigure} 
    \begin{subfigure}[b]{0.24\textwidth}
    \centering
    \includegraphics[width=\columnwidth]{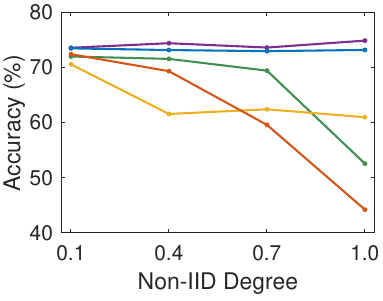}
    \subcaption{T2 (DYN-OPT)}
    \end{subfigure} 
    \begin{subfigure}[b]{0.24\textwidth}
    \centering
    \includegraphics[width=\columnwidth]{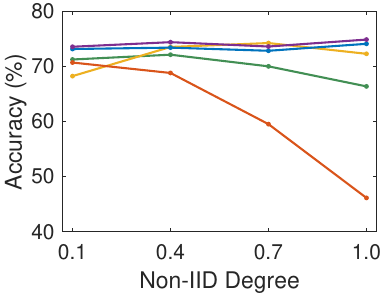}
    \subcaption{T3 (LIE)}
    \end{subfigure} 
    \begin{subfigure}[b]{0.24\textwidth}
    \centering
    \includegraphics[width=\columnwidth]{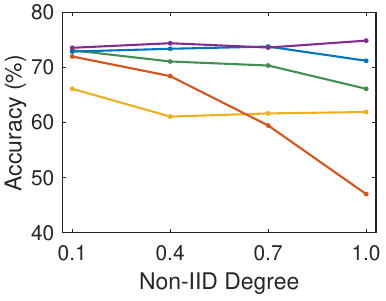}
    \subcaption{T4 (LIE)}
    \end{subfigure}
\centering
\caption{Impact of Fraction of Malicious Clients (FEMNIST, (a)-(d)) and Non-IID Degree (CIFAR10, (e)-(h)) on Accuracy against MPAs (STAT-OPT (Tmean), Min-Sum (\textit{sgn}), DYN-OPT (Tmean), LIE (\textit{sgn})) and Threat Models (T1-T4) }
\label{fig:5_ablation}
\end{figure*}

Training the global model with non-IID setting in FL is a challenge that has been studied vigorously recently and as the degree of non-IID increases, it becomes more difficult to train a global model with high accuracy. Sub-figures of (e) to (h) (i.e., second row) in~\cref{fig:5_ablation} illustrate the impact of non-IID degree on FLGuard compared with other byzantine-robust AGRs with CIFAR-10. With FedAvg as the baseline, we observed that other defenses fail to maintain their robustness against MPAs as the degree increases. FLGuard is able to withstand poisoning attacks even when the data is non-IID to a great extent as the performance of the global model using FLGuard is consistent and similar to that of FedAvg.

%% file: Tables/2_type_1_2.tex
\resizebox{\columnwidth}{!}{
    \Large
    \begin{tabular}{|c|c||c||C{1.35cm}C{1.35cm}|C{1.35cm}C{1.35cm}|C{1.35cm}C{1.35cm}||C{1.35cm}C{1.35cm}|C{1.35cm}C{1.35cm}|C{1.35cm}C{1.35cm}|}

    \hline
    \multirow{3}{*}{\makecell{Dataset \\ (Distr.)}} & 
    \multirow{3}{*}{AGR} & 
    \multirow{3}{*}{\makecell{No \\Attack}} & 
    \multicolumn{6}{c||}{Type-1} &
    \multicolumn{6}{c|}{Type-2} \\
\cline{4-15}

& 
& 
&
\multicolumn{2}{c|}{STAT-OPT} & 
\multicolumn{2}{c|}{DYN-OPT} & 
\multicolumn{2}{c||}{Adaptive} & 
\multicolumn{2}{c|}{STAT-OPT} & 
\multicolumn{2}{c|}{DYN-OPT} & 
\multicolumn{2}{c|}{Adaptive} \\

& 
& 
&
{Krum}  & {TM} & 
{Krum}  & {TM} & 
{DnC}   & {FLG}   & 
{Krum}  & {TM} & 
{Krum}  & {TM} & 
{DnC} & {FLG} \\

\hline
\hline

\multirow{8}{*}{\makecell{MNIST\\-0.1 \\ (IID)}} &

TM&
96.98& 89.83& 96.16& 87.60& 90.30& 86.85& 96.33& 
86.67& 95.56&  87.30& 87.28&  96.90& 95.31\\

& 
MK& 96.37& 84.33& 95.86& 84.11& 96.45& 83.87& 94.68& 
83.95& 94.58& 83.83& 96.45& 96.61& 96.60\\
  
&
Bul& 95.92& 88.29& 93.65& 88.94& 95.19& 87.36& 96.10& 
86.67& 93.34& 90.71& 95.21& 94.56& 93.04\\

& 
DnC& 97.06& 93.43& 96.77& 96.83& \textbf{96.88}& 96.83& 95.54& 
91.72& 96.53& 96.69& 96.75& 97.00& 94.74\\

& 
FLT& 95.96& 94.72& 95.50& 95.27& 93.69& 95.54& 93.32& 
92.94& 95.25& 94.93& 93.73& 94.06& 93.41\\
&
SG& 97.20& 95.90& \textbf{96.85}& 92.63& 96.81& 93.32& 96.47&
91.56& \textbf{96.92}& 89.33& \textbf{96.96}& 97.22& 95.39\\

\rowcolor{Gray}
\cellcolor{white}
& 
\textbf{FLG}& \textbf{97.24}& \textbf{96.83}& 96.79& \textbf{96.85}& 96.83& \textbf{96.90}& \textbf{96.85}&
\textbf{96.85}& 96.85& \textbf{96.88}& 96.83& \textbf{97.26}& \textbf{97.06}\\

\hline
\hline

\multirow{8}{*}{\makecell{CIFAR\\10 \\ (IID)}} &

TM& 71.92& 58.16& 46.53& 63.82& 52.11& 69.32& 71.83& 
58.73& 44.48& 58.89& 53.33& 59.88& 71.39\\

& 
MK& 71.41& 31.82& 45.11& 41.94& 71.49& 51.58& 69.56&
41.66& 41.92& 49.49& 71.06& 68.06& 69.58\\
  
& 
Bul& 56.62& 31.84& 40.38& 35.86& 49.39& 41.96& 67.51&
33.12& 38.35&  36.47& 48.15&  50.89& 64.89\\
 
& 
DnC& 72.44& 72.73& 71.55& 64.94& 72.65& 47.28& 70.88& 
72.22& 72.85& 70.76& 71.96&  72.08& 70.84\\

& 
FLT& 70.74& 71.00& 56.98& 65.02& 70.80& 67.45& 71.47& 
70.27& 53.04& 71.29& 70.54&  71.06& 68.99\\
&
SG& 70.64& 66.72& 72.52& 68.63& 69.66& 70.21& 69.70&
60.88& 72.71& 68.71& 72.38& 70.86& 70.68\\

\rowcolor{Gray}
\cellcolor{white}
& 
\textbf{FLG}& \textbf{72.73}& \textbf{73.44}& \textbf{72.71}& \textbf{73.86}& \textbf{73.60}& \textbf{72.48}& \textbf{72.02}& 
\textbf{73.19}& \textbf{72.99}& \textbf{73.15}& \textbf{73.44}& \textbf{73.30}& \textbf{72.36}\\

\hline
\hline

\multirow{8}{*}{\makecell{MNIST\\-0.5 \\ (Non\\-IID)}} & 

TM& 95.96& 87.44& 95.19& 87.95& 90.00& 88.23& 96.33&
83.10& 93.45& 84.19& 87.18& 96.12& 95.31\\

& 
MK& 96.19& 56.43& 94.93& 52.72& 96.33&
62.74& 94.68& 59.01& 93.26& 58.22& 96.33& 96.08& 94.60\\
& 
Bul& 94.38& 86.89& 94.93& 88.43& 93.75&	88.13& 96.10&
85.17& 94.62& 88.11& 93.65& 94.46& 93.04\\
& 
DnC& 96.57& 85.35& \textbf{96.92}& 91.40&	96.81&	93.38&	95.54&	85.47&	96.75&	87.36&	96.85&	96.63&	94.74\\
& 
FLT& 95.47& 88.74&	93.65&	93.87&	89.02&	93.95&	93.32&	88.15&	94.03&	93.45&	88.35&	93.99&	93.41 \\
&
SG& \textbf{96.94}& 89.14&	95.84&	92.65&	96.94&	91.74&	\textbf{96.47}&	93.93&	96.94&	92.78&	96.73&	\textbf{97.02}&	95.39\\

\rowcolor{Gray}
\cellcolor{white}
& 
\textbf{FLG}& 96.79& \textbf{96.81}&	96.49&	\textbf{96.83}&	\textbf{96.96}&	\textbf{96.71}&	96.06&	\textbf{96.67}&	\textbf{96.96}&	\textbf{96.67}&	\textbf{96.96}&	97.00&	\textbf{96.71}\\

\hline
\hline

\multirow{8}{*}{\makecell{FE\\MNIST \\ (Non\\-IID)}} &

TM&  80.62& 56.78& 76.10& 62.52& 63.69& 49.44& 80.06& 
73.98& 76.73& 74.28& 75.68& 78.84& 81.11\\
& 
MK& 83.69& 5.05& 78.60& 5.02& 83.71& 4.87& 81.00& 
5.67& 78.13&  4.85& 83.74&  8.81& 81.34\\
& 
Bul& 69.89& 53.41& 72.77& 53.13& 66.74& 53.47& 75.95& 
48.89& 74.98&  53.53& 67.20&  56.50& 79.63\\
& 
DnC& 83.87& 6.97& 82.76& 4.85& 84.03& 4.87& 80.71& 
5.26& 81.06& 4.98& 83.63& 26.79& 80.65\\
& 
FLT& 81.83& 4.60& 83.30& 35.79& 4.58& 52.83& 80.07& 
4.68& 83.61& 36.88& 4.46& 5.09& 79.98\\
&
SG& 83.56& 80.37& \textbf{84.19}& 10.12& 83.58& 8.87& \textbf{82.15}& 
8.75& 83.96& 8.77& 83.40& 77.64& 81.73\\
    
\rowcolor{Gray}
\cellcolor{white}
& 
\textbf{FLG}& \textbf{84.74}& \textbf{84.14}& 83.80& \textbf{84.30}& \textbf{84.19}& \textbf{83.22}& 81.86& 
\textbf{83.12}& \textbf{84.02}& \textbf{82.11}& \textbf{83.94}& \textbf{81.51}& \textbf{83.44}\\
    \hline
    \end{tabular}%
}

%% file: Tables/3_type_3_4.tex
\resizebox{\columnwidth}{!}{
    \Large
    \begin{tabular}{|c|c||C{1.35cm}C{1.35cm}|C{1.35cm}C{1.35cm}|C{1.35cm}C{1.35cm}||C{1.35cm}C{1.35cm}|C{1.35cm}C{1.35cm}|C{1.35cm}C{1.35cm}|C{1.35cm}|}

    \hline
    \multirow{3}{*}{\makecell{Dataset \\ (Distr.)}} & 
    \multirow{3}{*}{AGR} & 
    \multicolumn{6}{c||}{Type-3} &
    \multicolumn{7}{c|}{Type-4} \\
    \cline{3-15}

& 
&
\multicolumn{2}{c|}{LIE} & 
\multicolumn{2}{c|}{Min-Max} & 
\multicolumn{2}{c||}{Min-Sum} & 
\multicolumn{2}{c|}{LIE} & 
\multicolumn{2}{c|}{Min-Max} & 
\multicolumn{2}{c|}{Min-Sum} &
\multirow{2}{*}{SF}\\

& 
&
\textit{uv} & \textit{sgn} & 
\textit{uv} & \textit{sgn} & 
\textit{uv} & \textit{sgn} & 
\textit{uv} & \textit{sgn} & 
\textit{uv} & \textit{sgn} & 
\textit{uv} & \textit{sgn} &
\multicolumn{1}{c|}{}\\

\hline
\hline

\multirow{8}{*}{\makecell{MNIST\\-0.1 \\ (IID)}} &

TM& 
96.90&	\textbf{97.00}&	91.80&	87.13&	94.62&	89.51&	
96.49&	96.51&	88.25&	86.40&	90.50&	90.10& 88.72\\

& 
MK&
\textbf{97.02}&	96.45&	87.78&	84.13&	90.26&	85.21&	
96.31&	96.45&	86.79&	83.77&	89.53&	85.02& 87.87\\

&
Bul&
96.90&	96.12&	89.63&	91.72&	92.23&	83.64&
95.45&	96.06&	86.99&	91.42&	92.21&	85.98& 92.23\\

& 
DnC&
96.85&	96.94&	\textbf{96.88}&	96.79&	94.99&	86.53&	96.88&	96.77&	96.81&	96.67&	93.10&	86.83&  96.85\\

& 
FLT&
95.35&	95.70&	93.97&	95.35&	95.43&	95.50&	95.74&	95.86&	93.41&	95.23&	93.41&	95.11&  89.65\\

&
SG&
96.98&	96.81&	96.12&	91.68&	\textbf{96.69}&	95.52&	\textbf{97.38}&	\textbf{96.94}&	95.58&	90.28&	96.33&	93.43&  96.61\\

\rowcolor{Gray}
\cellcolor{white}
& 
FLG&
96.90&	96.83&	96.85&	\textbf{96.83}&	96.53&	\textbf{96.88}&	96.83&	96.83&	\textbf{96.83}&	\textbf{96.88}&	\textbf{96.67}&	\textbf{96.85}&  \textbf{96.90}\\

\hline
\hline

\multirow{8}{*}{\makecell{CIFAR\\10 \\ (IID)}} &

TM& 
72.65&	71.61&	57.65&	60.65&	69.36&	69.07&	73.42&	\textbf{75.28}&	64.85&	59.33&	68.34&	64.67&  57.73\\

& 
MK& 
71.31&	71.55&	65.32&	44.18&	65.69&	56.39&	70.64&	70.35&	59.21&	43.32&	64.02&	46.61&  62.62\\

& 
Bul&
73.42&	53.86&	38.80&	37.58&	45.21&	41.07&	64.96&	54.44&	39.31&	38.70&	40.28&	42.67&  48.54\\

& 
DnC&
72.24&	71.23&	71.61&	71.29&	70.94&	55.05&	72.24&	73.09&	71.83&	71.57&	71.49&	60.13&  \textbf{73.54}\\

& 
FLT&
72.24&	68.22&	71.29&	70.54&	72.24&	57.57&	71.00&	66.11&	70.27&	71.45&	70.45&	57.35&  69.83\\

&
SG&
72.50&	70.68&	60.11&	69.99&	70.05&	69.87&	72.81&	71.96&	58.85&	68.43&	69.95&	68.51&  53.06\\

\rowcolor{Gray}
\cellcolor{white}
& 
\textbf{FLG}&
\textbf{73.60}&	\textbf{73.13}&	\textbf{72.95}&	\textbf{73.50}&	\textbf{72.38}&	\textbf{72.50}&	\textbf{74.25}&	72.85&	\textbf{72.87}&	\textbf{72.97}&	\textbf{72.06}&	\textbf{72.87}&  72.06\\

\hline
\hline

\multirow{8}{*}{\makecell{MNIST \\ -0.5 \\ (Non\\-IID)}} &

TM& 97.04& 96.88& 92.55& 87.36& 95.37& 88.86& 
96.23& 96.29& 91.80& 83.52& 95.21& 86.97& 86.93\\
& 
MK& 97.08& 96.33& 89.35& 54.20& 95.45& 65.44& 
95.88& 96.33&  88.31& 56.55& 93.81& 64.27& 93.63\\
& 
Bul& 96.73& 95.03& 86.49& 89.87& 84.64& 80.64& 
95.01& 95.01& 86.32& 88.84& 94.80& 80.88& 89.91\\
& 
DnC& 97.10& 96.88& 96.53& 96.73& 95.35& 80.22& 
96.59& 96.77& \textbf{95.03}& \textbf{96.79}& 93.51& 79.00& 96.63\\
& 
FLT& 94.50& 94.18& 88.68& 93.28& 91.70& 76.22& 
94.30& 94.30& 90.16& 92.98& 90.10& 82.47& 91.34\\
&
SG& \textbf{97.18}& 96.92& 95.03& 93.69& \textbf{96.33}& 92.98& 
96.85& 96.69& 92.31& 92.13& 95.25& 89.79& 96.29\\

\rowcolor{Gray}
\cellcolor{white}
& 
FLG& 96.90& \textbf{96.96}& \textbf{96.59}& \textbf{96.96}& 96.27& \textbf{96.33}&
\textbf{97.10}& \textbf{96.96}& 94.40& 96.63& \textbf{95.39}&  \textbf{96.47}& \textbf{96.90}\\

\hline
\hline

\multirow{8}{*}{\makecell{FE\\MNIST \\ (Non\\-IID)}} &

TM& 
83.12& 83.95& 72.09& 57.64& 81.26& 64.11&
82.21& 83.17& 73.62& 71.05& 80.49& 72.24& 79.73\\

& 
MK&  83.86& 72.30& 80.18& 4.85& 82.90& 9.33& 
83.68& \textbf{83.80}& 78.13& 4.87& 82.79& 11.25& 78.23\\

& 
Bul& 
82.60&	71.41&	58.29&	61.21&	72.34&	34.50&	80.86&	72.28&	57.43&	60.39&	73.09&	45.37&  68.70\\

& 
DnC&
83.93&	83.59&	83.34&	44.11&	83.36&	5.69&	83.84&	83.63&	80.93&	5.25&	83.51&	5.42&   81.44\\

& 
FLT&
\textbf{84.92}&	81.97&	4.64&	59.68&	76.16&	58.83&	83.66&	4.85&	6.64&	6.17&	5.63&	6.40&   14.27\\

&
SG&
83.90&	83.56&	80.09&	8.73&	83.58&	76.80&	83.79&	83.74&	80.10&	8.13&	83.27&	10.80&  78.43\\
    
\rowcolor{Gray}
\cellcolor{white}

& 
\textbf{FLG}&
84.32& \textbf{84.04}& \textbf{84.07}& \textbf{84.39}& \textbf{84.19}& \textbf{82.62}& \textbf{83.90}& 82.41& \textbf{83.47}& \textbf{82.63}& \textbf{84.08}& \textbf{83.53}& \textbf{83.79}\\
    \hline
    \end{tabular}%
}

%% file: Tables/4_type_5.tex
\begin{tabular}{|c||cc|cc|cc|cc|cc|cc|cc|cc|}
\hline
\multirow{2}{*}{AGR}& 
\multicolumn{8}{c|}{Type-5} \\
\cline{2-9}
&
SLF & DLF & SLF & DLF & SLF & DLF & SLF & DLF \\
\hline
\hline
&
\multicolumn{2}{c|}{MNIST-0.1} &
\multicolumn{2}{c|}{CIFAR-10} &
\multicolumn{2}{c|}{MNIST-0.5} &
\multicolumn{2}{c|}{FEMNIST} \\
\cline{2-9}
TM&
95.50&	96.02&
62.72&	47.93&
94.81&	95.78&
80.39&	75.21\\

MK&
96.45&	96.45&
71.21&	70.58&
96.25&	96.31&
83.47&	84.37\\

Bul& 
95.78&	95.68&
46.49&	53.31&
94.30&	94.56&
68.39&	72.17\\

DnC& 
96.67&	\textbf{96.83}&
71.23&	72.18&
96.67&	96.33&
84.15&	83.52\\

FLT& 
95.80&	95.64&
68.97&	70.19&
94.89&	95.29&
79.34&	81.84\\

SG& 
96.81&	96.81&
\textbf{71.67}&	72.48&
\textbf{96.77}&	\textbf{96.90}&
85.08&	84.55\\

\rowcolor{Gray}
FLG& 
\textbf{96.83}&	\textbf{96.83}&
68.06&	\textbf{72.63}&
96.16&	95.72&
\textbf{85.18}&	\textbf{84.67}\\

\hline

\end{tabular}%

%% file: Latex/6_conclusion.tex
FLGuard is a novel byzantine-robust FL technique that employs contrastive learning to formulate the challenge of detecting and removing malicious clients as a self-supervised learning problem. Thus, unlike other byzantine-robust FL schemes, FLGuard operates without prior knowledge about FL schemes or auxiliary datasets. The contrastive models are trained.
Specifically, it adopts an ensemble of contrastive models trained with different dimension reduction methods and a clustering method to collect local model updates from only benign clients.
We conducted an extensive evaluation with various experimental settings, including IID and non-IID datasets and various poisoning attacks (i.e., MPAs and DPAs) under five different threat models. The results demonstrate that FLGuard outperforms existing defense mechanisms by achieving three defender's objectives: fidelity, robustness, and efficiency. In particular, FLGuard has improved the accuracy of the global model with non-IID datasets by a huge margin. 
Moreover, our ablation study reveals that the performance of FLGuard is maintained even when a large fraction of clients are compromised and the degree of non-IID is high.
We further note that FLGuard can be further improved by increasing the number of negative pairs used in contrastive learning. Also, we observed cases where dimension-wise filtering performed better than vector-wise defenses. Hence, we can devise a hybrid approach to combine the two for a stronger byzantine-robust FL scheme which we leave as future work.

%% file: Latex/7_appendix.tex
\appendix
\section{Appendix}
\subsection{FLGuard Algorithms and Parameter settings for experiments} \label{appendixA}
\cref{alg:LocalUpdate} illustrates the algorithm for updating the local model and returning the local updates for the server to gather.~\cref{alg:Filtering} depicts the filtering process of FLGuard through the usage of contrastive models and clustering. 
We set the parameter $e{=}1.5$ for DnC and collect the root dataset as the same distribution as the overall training dataset for FLTrust. For FLGuard, we set learning rate${=} 0.001$, epoch${=} 5$, batch size${=} 32$ and \textit{k}${=}5$ for all experiments.

\begin{algorithm}
\caption{LocalUpdate($w$, $I$, $D$, $b$, $F$, $\alpha$)}
\begin{algorithmic}[1]
    \label{alg:LocalUpdate}
    \STATE $w_0 \gets w$ 
    \FOR{each local iteration $i = 1,2,...,I$} 
        \STATE each mini-batch $b$ sampled from $D$
        \STATE $w_i \gets w_{i-1} - \alpha \cdot \frac{\partial F(b,w_{i-1})}{\partial w_{i-1}} $ 
    \ENDFOR
    \RETURN $w_I - w$
\end{algorithmic}
\end{algorithm}

\begin{algorithm}
\caption{Filtering($G_{lv}, G_{rd}, Contrastive Models$)}
\begin{algorithmic}[1]
    \label{alg:Filtering}
    \STATE $h_{lv}, h_{rd} \gets \textbf{ContrastiveModels} (G_{lv}, G_{rd})$
    \STATE $h'_{lv}, h'_{rd} \gets \textbf{PCA}(h_{lv}, h_{rd}, var) $
    \STATE $C_{lv}, C_{rd} \gets \textbf{Clustering}(h'_{lv}, h'_{rd})$

    \STATE $C_{good} \gets C_{lv} \cap C_{rd}$ \COMMENT{Ensemble Method}
    \RETURN $C_{good}$
\end{algorithmic}
\end{algorithm}

\begin{table}
    \centering
    \caption{FL parameter settings for various datasets. $\eta$ for CIFAR10 is with weight decay of 5E-4 and momentum of 0.9} 
    \input{Tables/5_param_setting}
    \label{tbl:5_param_setting}
\end{table}

\subsection{Details of Evaluated Attacks} \label{appendixB}
\textbf{Optimization Approaches}~\cite{shejwalkar2022back} consist of two types: static and dynamic. Such approaches assume that the aggregation rule for FL is known to the adversaries. 
\textbf{Static optimization (STAT-OPT})~\cite{fang2020local} searches for a sub-optimal $\gamma$ in generating malicious local updates $g_m$. The objective is to bypass the aggregation rule (i.e., classified as benign) and enable the malicious local updates ($g_m$) to participate in training the global model. 
\textbf{Dynamic approach (DYN-OPT})~\cite{shejwalkar2021manipulating} operates under the same constraints but differs in that DYN-OPT dynamically finds the largest $\gamma$ for the specific AGR. Therefore, DYN-OPT is a stronger attack that circumvents the byzantine-robust AGR better. 
\textbf{Adaptive Attacks}~\cite{shejwalkar2021manipulating} against DnC and FLGuard are reported separately. 
The adaptive attack against DnC operates similarly to DYN-OPT in that the largest $\gamma$ for a specific known AGR is obtained. However, FLGuard updates the contrastive models every $k$ FL rounds which makes designing the adaptive attack more challenging. In other words, the mechanism of AGR used to design the adaptive attacks is replaced with new contrastive models that project the local update to a different representation map. Therefore, in designing the adaptive attack against FLGuard, we assume that the adversaries can obtain the contrastive models every $k$ FL rounds and use DYN-OPT to find the largest $\gamma$.
\textbf{Little Is Enough (LIE)}~\cite{baruch2019little} crafted malicious updates that are within the population variance but yet cause byzantine failure in FL based on an observation that the practical variance of local model updates is high. LIE successfully degrades the accuracy of the global model with a very small $\gamma$ to bypass the previous defense mechanisms.
\textbf{Min-Max}~\cite{shejwalkar2021manipulating} aims to place malicious model update ($g_m$) close to the benign ones ($g_b$) by finding a suitable $\gamma$. Hence, the maximum distance of $g_m$ to any other $g$ is bounded by the maximum distance between two $g_b$. 
\textbf{Min-Sum}~\cite{shejwalkar2021manipulating} operates similar to Min-Max but with different upper bound constriant in finding $\gamma$. Min-Sum bounds the sum of squared distances between $g_m$ and all other $g_b$ by the sum of squared distances of any two $g_b$. 
\textbf{Sign-Flip Attack (SF)} reveres the local updates in the opposite direction without adjusting the size. This is a special case of the reversed gradient attack~\cite{rajput2019detox}.
\textbf{Label-Flip attack} is one of the data poisoning attacks. The malicious local updates are generated using data with flipped labels. Static Label-Flip (SLF)~\cite{fang2020local,cao2020fltrust} flips the original label of data $c$ to $n{-}1{-}c$, where $n$ is the total number of labels. Dynamic Label-Flip (DLF)~\cite{shejwalkar2022back} trains a surrogate model with the available benign data and flips the original label to the least probable label output from the surrogate model.

%% file: Tables/5_param_setting.tex
\begin{tabular}{|c|c|c|c|c|}
\cline{1-5}
&
Details & 
MNIST&
CIFAR-10 &
FEMNIST \\ 
\hline
\hline


$R$ & \# FL Rounds & 300 & 2,000 & 1,500 \\ 
\cline{1-5}

$N$ & \# Total Clients & 100 & 50 & 3,400 \\ 
\cline{1-5}

$M$ & \# Malicious Clients & 20 & 10 & 680  \\ 
\cline{1-5}


$P$ & \# Participating Clients & \multicolumn{2}{c|}{$N$} & 60 \\ 
\cline{1-5}

$b$ & Batch Size & 250 & 32 & 250 \\ 
\cline{1-5}

$\eta$ & Global Learning Rate & 0.001& 0.01 & 0.001 \\
\cline{1-5}


$Opt$ & Optimizer & Adam & SGD & Adam\\
\cline{1-5}
$Arch$ & Global Model Architecture &  FCN & ResNet-14 & ConvNet \\
\cline{1-5}
\end{tabular}